\documentclass[letterpaper]{article} 
\usepackage{aaai2026}  
\usepackage{times}  
\usepackage{helvet}  
\usepackage{courier}  
\usepackage[hyphens]{url}  
\usepackage{graphicx} 
\usepackage{natbib}  
\usepackage{caption} 
\usepackage{algorithm}
\usepackage{algorithmic}

\usepackage{newfloat}
\usepackage{listings}
\DeclareCaptionStyle{ruled}{labelfont=normalfont,labelsep=colon,strut=off} 
\floatstyle{ruled}
\newfloat{listing}{tb}{lst}{}
\floatname{listing}{Listing}
\title{\m{}: Interaction-Centric Video Generation via Mask Trajectories}
\author{Gen Li\textsuperscript{\rm1,3},
Bo Zhao\textsuperscript{\rm2\footnotemark[1]},
Jianfei Yang\textsuperscript{\rm1\thanks{Corresponding authors.}},
Laura Sevilla-Lara\textsuperscript{\rm3}}

\affiliations{
\textsuperscript{\rm 1}Nanyang Technological University\\
\textsuperscript{\rm 2}Shanghai Jiao Tong University\\
\textsuperscript{\rm 3}University of Edinburgh\\
\{gen.li, jianfei.yang\}@ntu.ed.sg, bo.zhao@sjtu.edu.cn
}

\usepackage{bibentry}

\usepackage{amsfonts}
\usepackage{amsmath}
\usepackage{booktabs}
\usepackage{multirow}
\usepackage{makecell}
\usepackage[capitalize]{cleveref}
\crefname{section}{Sec.}{Secs.}
\Crefname{section}{Section}{Sections}
\Crefname{table}{Table}{Tables}
\crefname{table}{Tab.}{Tabs.}
\newcommand{\m}[1]{Mask2IV}
\newcommand{\p}[1]{interaction-centric}

\begin{document}

\maketitle

\begin{abstract}


Generating interaction-centric videos, such as those depicting humans or robots interacting with objects, is crucial for embodied intelligence, as they provide rich and diverse visual priors for robot learning, manipulation policy training, and affordance reasoning.
However, existing methods often struggle to model such complex and dynamic interactions. While recent studies show that masks can serve as effective control signals and enhance generation quality, obtaining dense and precise mask annotations remains a major challenge for real-world use.
To overcome this limitation, we introduce \m{}, a novel framework specifically designed for interaction-centric video generation. It adopts a decoupled two-stage pipeline that first predicts plausible motion trajectories for both actor and object, then generates a video conditioned on these trajectories.
This design eliminates the need for dense mask inputs from users while preserving the flexibility to manipulate the interaction process.
Furthermore, \m{} supports versatile and intuitive control, allowing users to specify the target object of interaction and guide the motion trajectory through action descriptions or spatial position cues.
To support systematic training and evaluation, we curate two benchmarks covering diverse action and object categories across both human-object interaction and robotic manipulation scenarios.
Extensive experiments demonstrate that our method achieves superior visual realism and controllability compared to existing baselines.
\end{abstract}

\begin{links}
    \link{Project page}{https://reagan1311.github.io/mask2iv}
\end{links}

\begin{figure}[t]
  \centering
  \includegraphics[width=.474\textwidth]{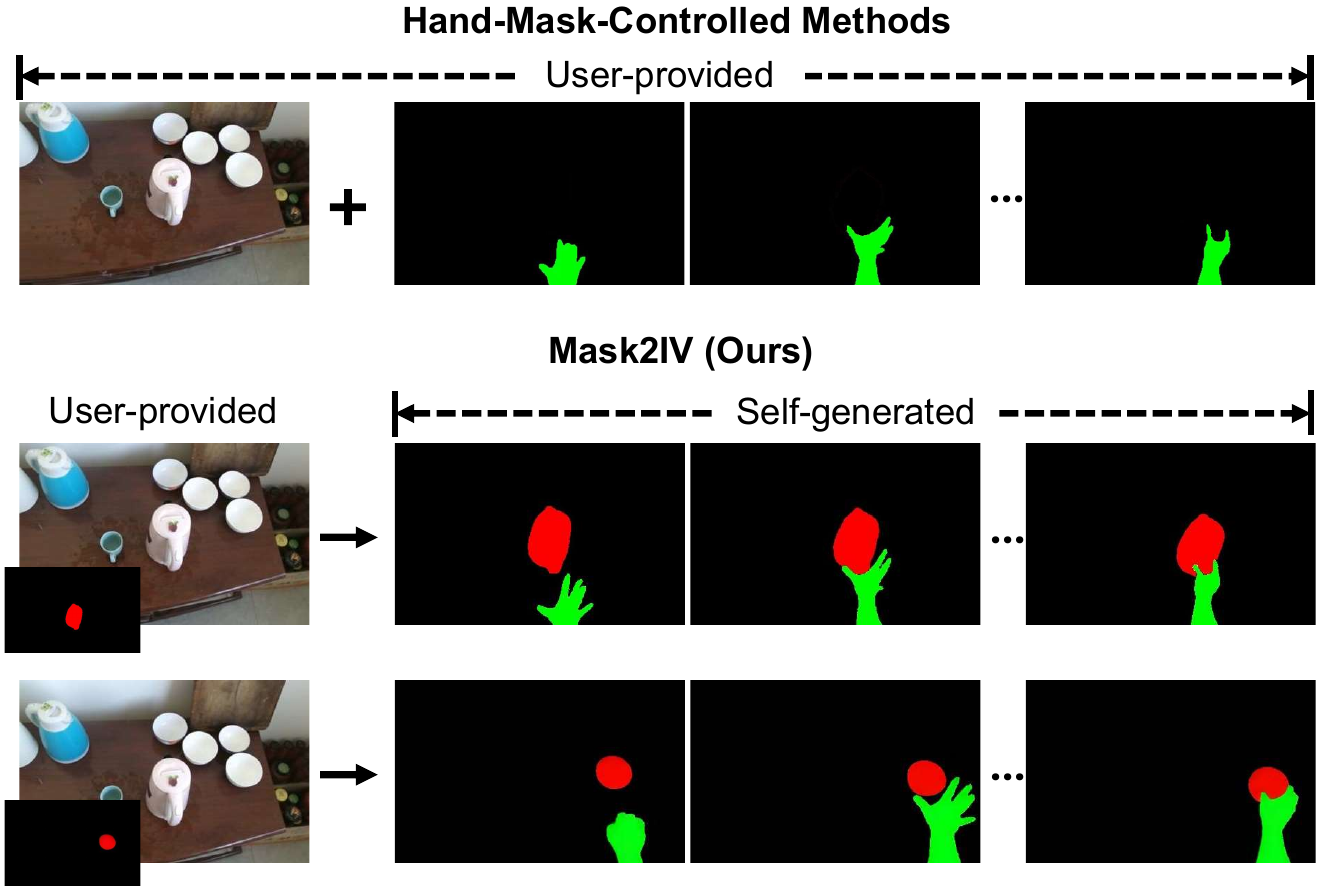}
  \caption{\textbf{Comparison on control signal acquisition}. Hand-mask-controlled video generation methods require users to provide dense hand mask sequences as input. 
  In contrast, \m{} autonomously generates trajectories for both hands and objects without manual annotation, and can adaptively produce different trajectories based on the specified object.}
  \label{fig:motivation}
\end{figure}

\begin{figure*}[t]
  \centering
  \includegraphics[width=.95\textwidth]{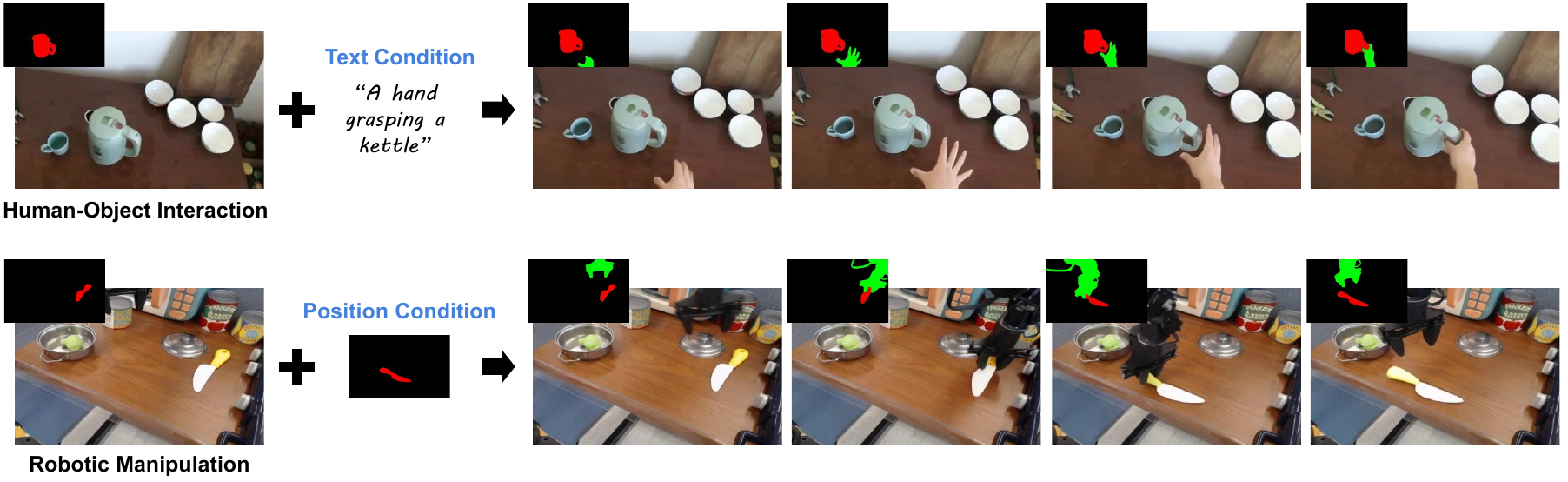}
  \caption{\textbf{\m{}} synthesizes videos of human hands or robot arms interacting with a specified object, indicated by an input mask. It first predicts a mask-based interaction trajectory (visualized in the top-left corner of each frame), and then generates the video guided by this trajectory. The generation process is conditioned on either a text prompt or a target position mask.}
  \label{fig:teaser}
\end{figure*}

\section{Introduction}
\label{sec:intro}

Interacting with objects is a fundamental aspect of daily human life: we actively engage with a wide variety of objects without explicit prior planning.
Our extensive experience with embodied interaction allows us to anticipate the appropriate hand pose, the trajectory required for a specific action, and the ideal placement of a target---even before physical contact occurs.
This remarkable imaginative ability stems from our rich prior knowledge of object dynamics and action understanding.
However, despite appearing effortless for humans, replicating this capacity in generative models or robotic systems remains a formidable challenge.
In the context of embodied AI, addressing this challenge is particularly valuable, as the ability to synthesize realistic, diverse, and controllable human-object or robot-object interaction sequences can provide powerful visual priors that facilitate tasks such as imitation learning~\cite{kareer2024egomimic,lepert2025phantom} and affordance learning~\cite{affgrasp,li2023locate}.
Moreover, high-fidelity modeling of these interactions is crucial for a wide range of applications, including augmented and virtual reality, robot learning, and motion planning.

Recent advances in diffusion models have substantially improved the generation of realistic and coherent visual content.
Nonetheless, current models often fall short in accurately capturing the complex and dynamic interactions between human hands or robotic manipulators and objects.
To this end, a growing body of research has emerged, with a focus on interaction-centric image and video generation.
Much of the existing work~\cite{lai2024lego,handi,egovid} targets egocentric views, where diverse human-object interaction behaviors are prevalent.
While promising results have been achieved, these methods typically rely on text-based conditioning, which is insufficient for modeling fine-grained interactions.
Specifically, they lack precise control over which object is being interacted with and where the hands are positioned.
To address this shortcoming, recent studies~\cite{coshand,interdyn} have proposed using hand masks as an explicit control signal for interaction modeling.
Given an input image of a hand interacting with an object, dense temporal hand masks are provided to guide the generation of subsequent frames, ensuring alignment with the intended interaction.
This hand mask control mechanism, while effective for generating plausible interactions, suffers from two critical limitations.
First, hand-mask conditioning is inherently impractical and user-unfriendly, as it necessitates frame-by-frame acquisition of precise hand masks.
Such masks can only be obtained by first recording or synthesizing the exact interaction video that one wishes to control. In practice, it is unrealistic to expect users to provide a complete sequence of temporal masks in advance.
Second, relying exclusively on hand masks constrains the scope of interaction modeling, which hinders both the precise specification of target objects for interaction and the accurate capture of fine-grained hand-object contact, particularly under camera motion.


To overcome these limitations, we propose~\m{} (Mask-to-Interaction Video generation), a two-stage framework that decomposes the video generation process into interaction trajectory generation and trajectory-conditioned video generation.
As illustrated in~\cref{fig:motivation}, unlike hand-mask-controlled methods such as InterDyn~\cite{interdyn}, our approach eliminates the need to provide dense mask sequences as control signals.
Instead, it predicts interaction trajectories based on an initial image and an object mask.
Moreover, \m{} produces trajectories of both actors and objects for control, which exhibits improved performance, especially in challenging egocentric and multi-object settings.
In addition, \m{} supports object-specific control by generating interaction trajectories conditioned on arbitrary target objects.

We demonstrate our model's capabilities across two key domains: human-object interaction and robotic manipulation.
To guide the generation of interaction trajectories, we explore two types of conditioning prompts, which are action descriptions and spatial position cues.
The former provides a high-level and intuitive specification of the intended action, while the latter enables precise and low-level control over object placement.
Examples generated by our approach are presented in~\cref{fig:teaser}, illustrating highly plausible and physically consistent interaction dynamics.
For training and evaluation, we curate dedicated benchmarks building on two widely used datasets: HOI4D~\cite{liu2022hoi4d} for human-object interaction and BridgeDataV2~\cite{walke2023bridgedata} for robotic manipulation.

To summarize, our main contributions are as follows:
\begin{itemize}
    \item We introduce the task of interaction-centric video generation, targeting realistic and controllable synthesis of human-object and robot-object interactions.
    \item We propose an innovative decoupled two-stage framework that not only improves the generation quality but also enables explicit control over the interaction target.
    \item We construct two dedicated benchmarks covering diverse interaction scenarios and conduct extensive experiments, demonstrating that our approach outperforms existing baselines in both controllability and visual fidelity.
\end{itemize}

\section{Related Work}
\label{sec:related_work}

\noindent \textbf{Human and Robot Interaction Synthesis.}
Synthesizing realistic and plausible human or robot interactions is a fundamental problem in computer vision and robotics.
A large body of work focuses on generating human interactions in 2D images, with different focuses on hand~\cite{hu2022hand,ye2023affordance,narasimhaswamy2024handiffuser,qin2024handcraft,park2024attentionhand,coshand,lai2024lego,souvcek2024genhowto} or the whole-body interaction~\cite{kulal2023putting,yang2024person,hoe2024interactdiffusion,xu2024semantic,jiang2024record,fang2024humanrefiner}.
With the recent progress in video generation, growing attention has turned toward modeling human-object interactions in dynamic scenes~\cite{furuta2024improving,xu2024anchorcrafter}.
EgoVid~\cite{egovid} addresses the data bottleneck by introducing a high-quality dataset curated for egocentric video generation.
HOI-Swap~\cite{xue2024hoi} and Re-HOLD~\cite{fan2025re} synthesize novel interactions by recombining hands and objects in videos.
InterDyn~\cite{interdyn} proposes a controllable generation framework using hand mask sequences, but requires dense annotations and is limited to close-up scenes.
In parallel, there has been increasing interest in robot interaction synthesis, particularly in the context of robot-object manipulation. 
Recent studies such as RoboGen~\cite{robogen}, Genesis~\cite{Genesis}, and TesserAct~\cite{zhen2025tesseract} leverage generative models to produce realistic and diverse robot interactions. These methods often target downstream applications such as training visuomotor policies or generating synthetic demonstrations.
In this work, we introduce the task of interaction-centric video generation and explore the synthesis of both human and robot interactions within a single framework.

\noindent \textbf{Controllable Video Generation.}
With the advent of diffusion models~\cite{ddpm}, video generation has seen significant advancements, enabling the synthesis of high-quality videos from text-based or image-based prompts.
One particular area of interest is controllable video generation, which aims to provide users with the ability to influence the generated content.
The success of conditional text-to-image generation methods, such as ControlNet~\cite{controlnet}, GLIGEN~\cite{gligen}, or T2I-Adapter~\cite{t2i-adapter}, have laid the groundwork for controllable video synthesis. 
Building on these advances, recent work has extended similar control mechanisms to the video generation.
A variety of control signals have been explored, including
bounding-boxes~\cite{wang2024boximator,luo2024ctrl}, masks~\cite{dai2023animateanything,yariv2025through,interdyn}, depth maps~\cite{chen2023control,liang2024movideo}, and optical flow~\cite{shi2024motion,liang2024movideo}.
Other efforts focus on camera-level constraints, using camera trajectories or point tracks to guide dynamic viewpoint changes~\cite{wang2024motionctrl,he2024cameractrl,geng2024motion}.
While dense control signals have enabled fine-grained control over layout and motion, they are often impractical and not user-friendly. Providing dense, temporally consistent annotations is especially challenging for dynamic interactions involving actors and objects.
To address this, we propose a two-stage approach that first predicts a sequence of masks depicting interaction trajectories, then uses these mask-based trajectories to guide subsequent video generation.
This design allows users to automatically obtain dense control signals, while retaining flexibility in controlling generated videos.

\section{Method}

\begin{figure*}[hbtp]
  \centering
  \includegraphics[width=0.935\textwidth]{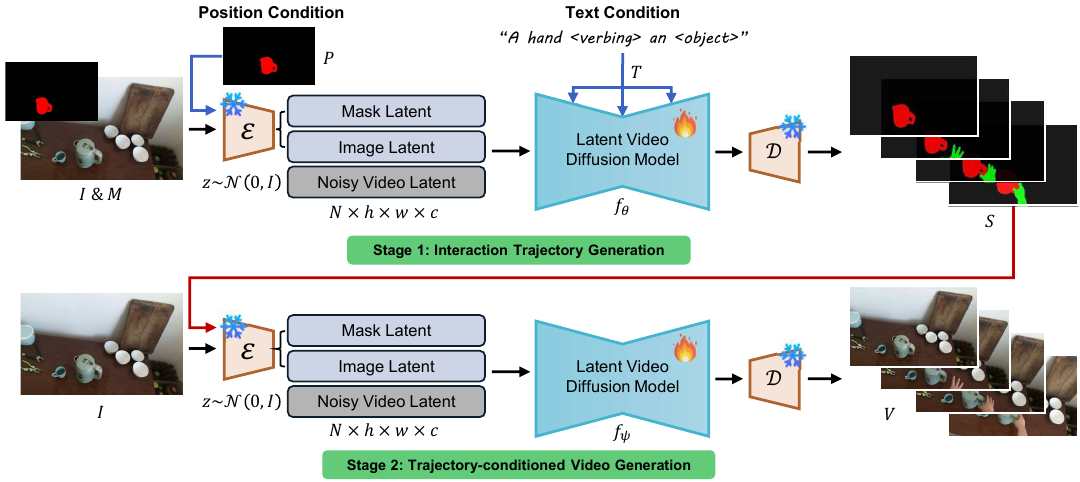}
  \caption{The framework of \m{}. It consists of two stages: Interaction Trajectory Generation and Trajectory-conditioned Video Generation. The first stage produces a mask-based interaction trajectory, while the second stage synthesizes a video conditioned on the predicted trajectory.}
  \label{fig:framework}
\end{figure*}

We focus on the problem of \p{} video generation, with the goal of enabling fine-grained control over the interaction process.
Specifically, we aim to satisfy the following three objectives:
(1) \textbf{Object Specification}. The object involved in the interaction can be explicitly designated.
(2) \textbf{Action Control}. The interaction can be guided using language descriptions of the intended action.
(3) \textbf{Target Localization}. The final position of the interacted object can be precisely controlled.

To achieve these goals, we propose \m{} that decomposes the generation process into two stages as illustrated in~\cref{fig:framework}:
\begin{itemize}
    \item \textbf{Interaction Trajectory Generation.}
    We simplify the problem by first generating the intermediate interaction trajectory, allowing the model to focus solely on motion dynamics without the complexity of appearance details.
    \item \textbf{Trajectory-conditioned Video Generation.}
    Leveraging the generated trajectory, we synthesize the final video while introducing specialized components to tackle the unique challenges of interaction generation.
\end{itemize}
This decomposition offers two key advantages: it mitigates the difficulty of directly generating intricate interaction dynamics, and provides greater flexibility for controlling the interaction process.
In this section, we first formalize the problem definition, then describe our two-stage framework in detail, and finally present the benchmark construction process for training and evaluation.

\subsection{Problem Formulation}
\label{sec:problem_formulation}

Given an RGB image $I\in \mathbb{R}^{H\times W\times 3}$ and a mask $M\in \mathbb{R}^{H\times W}$ that specifies the object of interest, the target is to generate a video $V\in \mathbb{R}^{N\times H\times W\times 3}$ consisting of $N$ frames that realistically portray dynamic interactions between the designated object and an actor (\textit{e.g.}, human hands or a robot arm).
To enable fine-grained control, we incorporate two complementary conditional inputs: (1) a text prompt $T$ that describes the intended action (\textit{e.g.}, a hand picking up a mug), 
and (2) a target position mask $P\in \mathbb{R}^{H\times W}$ that specifies the desired post-interaction placement of the object.
Our model flexibly adapts to either condition, ensuring that the generated video aligns with the described action or reflects the specified target location.
In contrast to prior work~\cite{interdyn}, our method eliminates the need for densely annotated actor masks during inference, greatly improving the practicability.

\subsection{Interaction Trajectory Generation}
\m{} tackles the interaction-centric video generation task with a two-stage framework. 
In the first stage, the model $f_\theta$ is trained to generate an interaction trajectory between the actor and the object, represented as a sequence of masks.
It takes as input the initial frame $I$, the object mask $M$, and a conditioning signal, either a text prompt $T$ or a position mask $P$, and outputs the interaction trajectory $S\in \mathbb{R}^{N\times H\times W\times 3}$.
Specifically, the input frame $I$ and object mask $M$ are encoded by the VAE encoder $\mathcal{E}$ into latent features $f_i\in \mathbb{R}^{h\times w\times 4}$ and $f_m\in \mathbb{R}^{h\times w\times 4}$, where $h=H/8$ and $w=W/8$. 
Since $\mathcal{E}$ requires a three-channel input, we first apply color encoding to the object mask $M$ to convert it into the RGB format. 
If the initial frame contains an actor, we use GroundedSAM~\cite{groundedsam} to perform segmentation and assign the actor a distinct color from that of the object, enabling the model to better differentiate their roles.
The resulting latent features are then combined along the channel dimension, duplicated $N$ times to match the video length, and concatenated with the noise latent $z\in \mathbb{R}^{N\times h\times w\times 4}$.
These combined features are fed into the latent video diffusion model, which, in conjunction with the VAE decoder $\mathcal{D}$, generates the interaction trajectory over time.
To preserve the motion priors, we freeze the temporal attention layer and fine-tune the remaining parameters of a pre-trained image-to-video diffusion model.

Under this setup, we explore two trajectory generation variants based on different types of conditioning: Text-conditioned Trajectory Generation (TT-Gen) and Position-conditioned Trajectory Generation (PT-Gen).
(1) TT-Gen leverages language guidance to shape the trajectory. It enables the model to distinguish between subtle interaction intents, such as pick up vs. put down, or push vs. pull. The text prompt $T$ is encoded using CLIP~\cite{clip} and injected into the model via cross-attention.
(2) PT-Gen focuses on precise spatial control of the object’s final position. The target position mask $P$ is encoded into a latent feature and inserted in the final frame’s slot, while the initial object mask latent is assigned to the first frame. All intermediate frames are filled with zero values, prompting the model to interpolate a coherent trajectory that transitions the object to the specified end position.

\subsection{Trajectory-conditioned Video Generation}
In the second stage, we fine-tune another model $f_\psi$ to generate the interaction video $V$, which is conditioned on the input image $I$ and the mask-based trajectory $S$ produced in the first stage.
Concretely, $S$ is firstly transformed into a latent feature tensor $f_s\in \mathbb{R}^{N\times h\times w\times 4}$ using the VAE encoder $\mathcal{E}$.
To enable control, $f_s$ is concatenated with the noise latent $z$ and the first-frame latent feature $f_i$, which is expanded $N$ times in the temporal dimension to match the video length.
The model is then trained with ground-truth mask sequences and infers using predicted masks from the first stage.

While the latent concatenation allows the generated video to follow the trajectory signal, the model lacks specific designs to handle the complex and dynamic interaction behaviors.
In particular, we observe two common issues:
(1) The synthesized videos are sensitive to variations in the mask trajectory.
(2) Regions around the boundary of the actor and the object, i.e., the contact area, is difficult to synthesize accurately.
We thus propose two designs to address these challenges.
First, we apply random perturbations to the trajectory $S$ to enhance the robustness of the generation process.
During training, with a probability $p$ (set to 0.2 in our case), we dilate or erode the mask using a kernel size randomly chosen from $\{3, 5, 7\}$.
Since different actors and objects may have masks of varying shapes, this operation encourages the model to generalize better, rather than strictly adhere to the exact shape of the input trajectory.
Second, we introduce a contact weighting loss to emphasize content in the interaction-rich regions.
Specifically, based on the hand mask $m_h$ and object mask $m_o$, a contact map $m_c$ is defined as:
\begin{equation}
m_c = \left( \delta(m_h) \cap m_o \right) \cup \left( m_h \cap \delta(m_o) \right),
\end{equation}
where $\delta(\cdot)$ denotes the dilation operation. The contact map is then used to reweight the diffusion objective, prioritizing contact regions:
\begin{align}
w &= (1 - m_c) + \lambda \cdot m_c,\\
\mathcal{L} &= \mathbb{E}_{z, S, \epsilon, t} \left[
\left\| w \odot \left( \epsilon - \epsilon_\theta(z, f_\psi(S), t) \right) \right\|_2^2
\right],
\end{align}
where $\lambda$ is a weighting factor denoting the importance of contact regions in the loss, $\epsilon_\theta(\cdot)$ represents the denoising network, and $t$ is the timestep in the forward process.

\subsection{Benchmark Construction}
To align with our focus on \p{} video generation, we target two representative and widely studied scenarios: Human-Object Interaction (HOI) and Robotic Manipulation.
Both scenarios require the modeling of fine-grained, temporally grounded interactions between actors and objects.
We explore text-conditioned trajectory generation for the HOI data, as language provides an intuitive and flexible way to specify human actions.
For robotic data, we adopt position-conditioned trajectory generation, since robotics tasks often involve pick-and-place actions that demand precise control over object placement.

We curate benchmarks with frame-level segmentation maps to support model training and evaluation.
For HOI, we select the HOI4D~\cite{liu2022hoi4d} dataset. It provides timestamped annotations for action start and end points, segmentation masks, and rich hand-object interactions, especially the grasp of diverse objects.
We crop the video clip using the timestamps and employ a text prompt template with the form ``a hand \{verbing\} an \{object\}".
To ensure sufficient interaction intensity, we compute a motion score based on the displacement of the hand and the object across each clip. 
Videos falling below the 5th percentile of this score distribution are filtered out to remove clips with minimal motion.
For robotic interactions, we adopt BridgeData V2~\cite{walke2023bridgedata}, a large dataset that captures robot manipulation across different environments with variation in objects, camera poses, and workspace configurations.
However, it does not provide segmentation masks. 
We thus utilize GroundingDINO~\cite{liu2024grounding} for object detection and SAM2~\cite{sam2} for video segmentation, targeting both the robot arm and the interacting object. 
Since the data is captured in cluttered scenes with multiple similar objects, we need to further identify the specific object involved in the interaction.
To this end, we compute the mean Intersection-over-Union (mIoU) of object masks across temporally spaced frames. 
Objects with low temporal mIoU are identified as those being manipulated, due to their changing shape and position over time.


\section{Experiments}

\subsection{Experimental Setup}
\noindent \textbf{Implementation Details.}
We build our method on DynamiCrafter~\cite{xing2024dynamicrafter}, one of the state-of-the-art image-to-video generation methods, and incorporate extra convolution channels to support encoding of the mask latent.
All experiments are conducted on two 80GB NVIDIA A100 GPUs. 
Each video consists of 16 frames, which are resized and cropped to a resolution of $320\times 512$.
We employ AdamW~\cite{adamw} optimizer with a learning rate of $1\times10^{-5}$ and a batch size of 8.
The weighting factor $\lambda$ is empirically set to 5 during training.
At inference, the DDIM sampler~\cite{ddim} is used with 50 timesteps to progressively denoise the latent representation and produce the output video.

\noindent \textbf{Metrics.}
We evaluate the generated videos from two key perspectives: generation quality and text-video alignment.
For generation quality, we employ Fr\'echet Video Distance (FVD)~\cite{fvd} to compute overall spatio-temporal similarity of videos, and use Peak Signal-to-Noise Ratio (PSNR), Structural Similarity Index Measure (SSIM)~\cite{ssim}, and Learned Perceptual Image Patch Similarity (LPIPS)~\cite{lpips} to measure the frame-level visual fidelity.
To measure text-video alignment, we use pretrained video-language models tailored to each dataset: EgoVLP~\cite{egovlp} for the HOI4D and ViCLIP~\cite{internvid} for the BridgeData V2.
We utilize the Text-to-Video Similarity (T2V-Sim) score to measure how closely the generated video matches the text prompt, computed as the cosine similarity between text and video embeddings.
Additionally, we compute the Video-to-Video Similarity (V2V-Sim) score, which is used to evaluate the alignment between generated and real videos in the embedding space.

\noindent \textbf{Baselines.}
We take two types of baseline methods for comparison.
The first is the pre-trained image-to-video diffusion model DynamiCrafter, which generates videos from a single input image and a text prompt, but lacks explicit control over the generated interaction.
We also include a fine-tuned version (DynamiCrafter-ft) to adapt the model to the curated datasets.
The second type consists of control-based method designed for interaction data generation, including CosHand and InterDyn.
However, these approaches require ground-truth control signals, which are not available in our setting and thus not directly comparable.
To ensure a fair and meaningful comparison, we adopt the same pseudo control signals produced by the first stage of our pipeline.
We also adapt their architectures where necessary to align with our setting.
Specifically, CosHand was originally proposed for image generation by concatenating the future mask latent with the noise latent.
To extend it for video generation, we concatenate mask latent features from all frames to provide temporal conditioning.
InterDyn is implemented similarly to its original form, except that we replace the ground-truth actor mask trajectories with our pseudo ones.


\begin{table*}[t]
    \small
    \centering
    \begin{tabular}{lccccccc}
    \toprule
    \textbf{Method} & \textbf{Pub.} & \textbf{FVD$\downarrow$} & \textbf{LPIPS$\downarrow$} & \textbf{PSNR$\uparrow$} & \textbf{SSIM$\uparrow$} & \textbf{V2V-Sim$\uparrow$} & \textbf{T2V-Sim$\uparrow$} \\
    \midrule
    DynamiCrafter & ECCV24 
    & 554.48 / 860.53  & 0.516 / 0.375 & 13.48 / 14.21 & 0.553 / 0.571 & 0.473 / 0.867 & 0.146 / 0.215 \\
    DynamiCrafter-ft & ECCV24 & 168.73 / 197.82 & 0.206 / 0.166 & 20.49 / 19.80 & 0.721 / 0.775 & 0.814 / 0.957 & 0.199 / \textbf{0.223} \\
    
    \midrule
    CosHand  & ECCV24 & 162.87 / 174.84 & 0.209 / 0.123 & 20.67 / 21.81 & 0.725 / 0.809 & 0.837 / 0.969 & 0.191 / 0.220 \\
    InterDyn & CVPR25 & 172.42 / 207.80 & 0.207 / 0.145 & 20.71 / 21.16 & 0.730 / 0.802 & 0.794 / 0.955 & 0.172 / 0.219 \\
    
    Mask2IV (Ours) & This Work & \textbf{149.68} / \textbf{155.73} & \textbf{0.178} / \textbf{0.111} & \textbf{21.48} / \textbf{22.30} & \textbf{0.741} / \textbf{0.815} & \textbf{0.847} / \textbf{0.971} & \textbf{0.200} / 0.220 \\
    \bottomrule
    \end{tabular}
    \caption{Quantitative comparisons on HOI4D / BridgeData V2. DynamiCrafter and its fine-tuned variant perform image-to-video generation without explicit control, while the remaining methods incorporate control signals through the use of masks.}
    \label{tab:comparison-hoi4d}
\end{table*}

\begin{figure*}[t]
  \centering
  \includegraphics[width=1.0\textwidth]{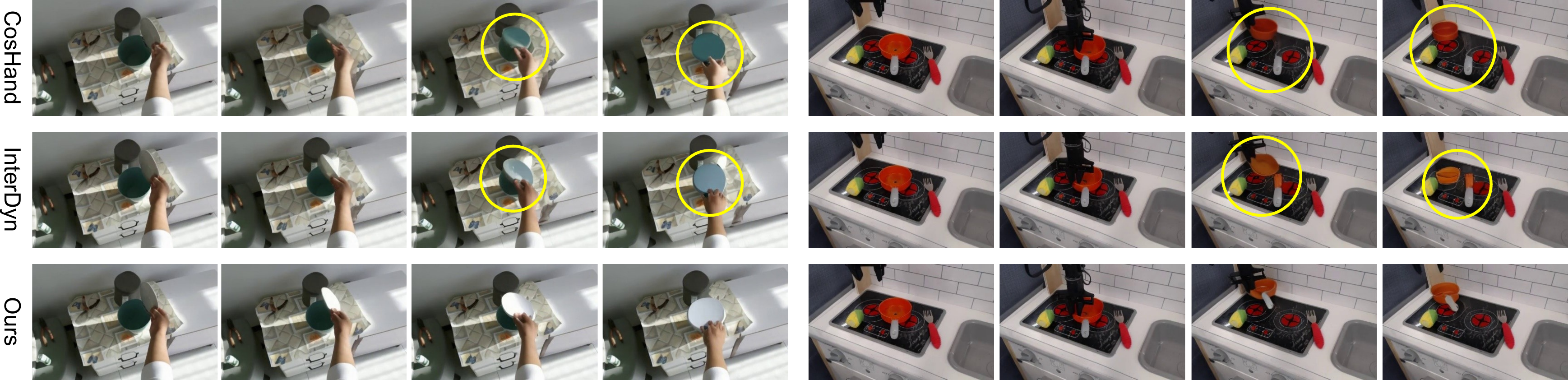}
  \caption{Qualitative comparison with CosHand and InterDyn. Generation artifacts are marked with yellow circles.}
  \label{fig:quali_analysi}
\end{figure*}

\begin{figure}[t]
  \centering
  \includegraphics[width=0.474\textwidth]{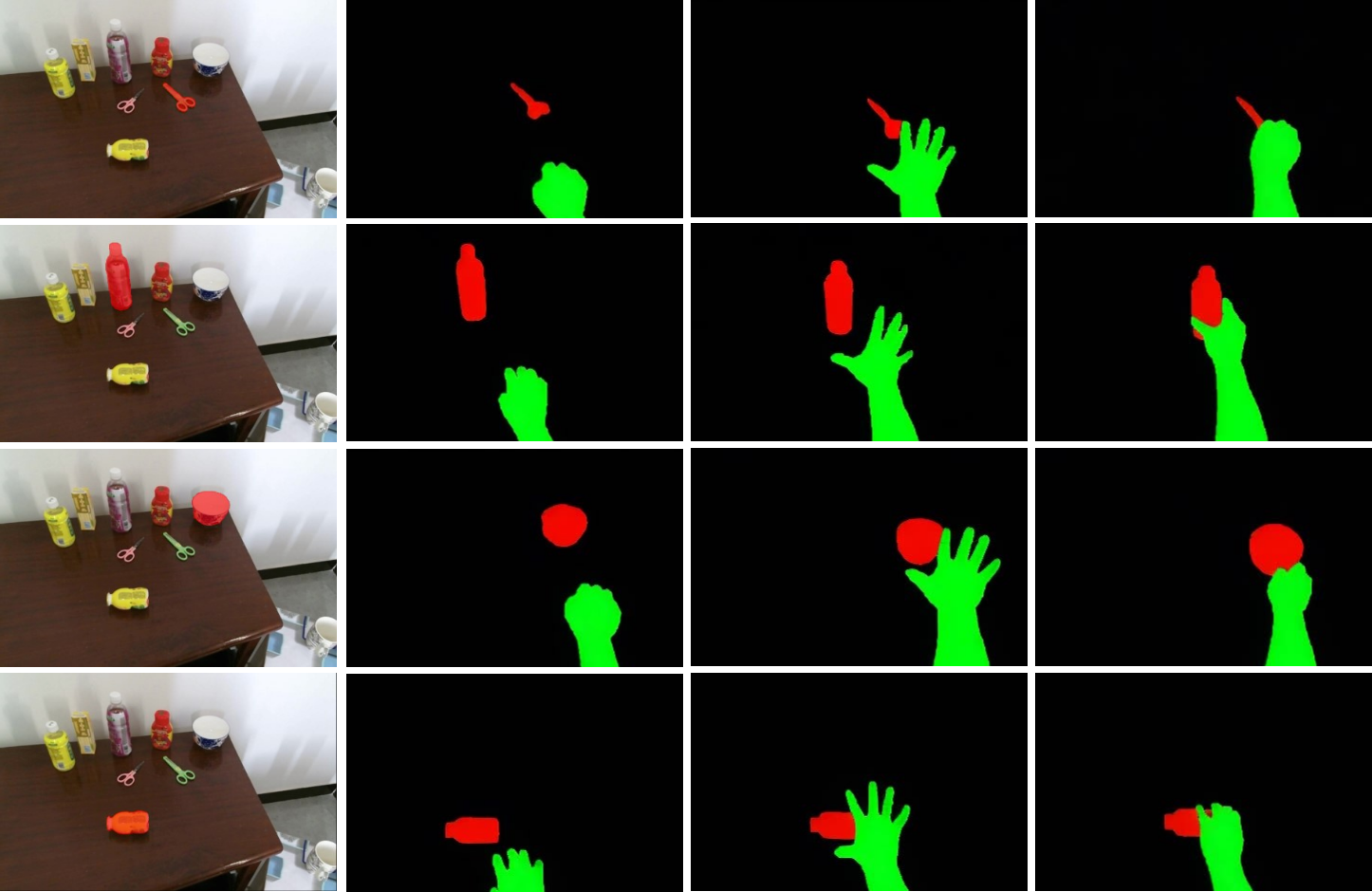}
  \caption{Qualitative analysis of interaction trajectory generation involving different objects. Target objects are highlighted in red in the initial image.}
  \label{fig:quali_analysis_obj}
\end{figure}

\begin{figure*}[htb]
  \centering
  \includegraphics[width=0.885\textwidth]{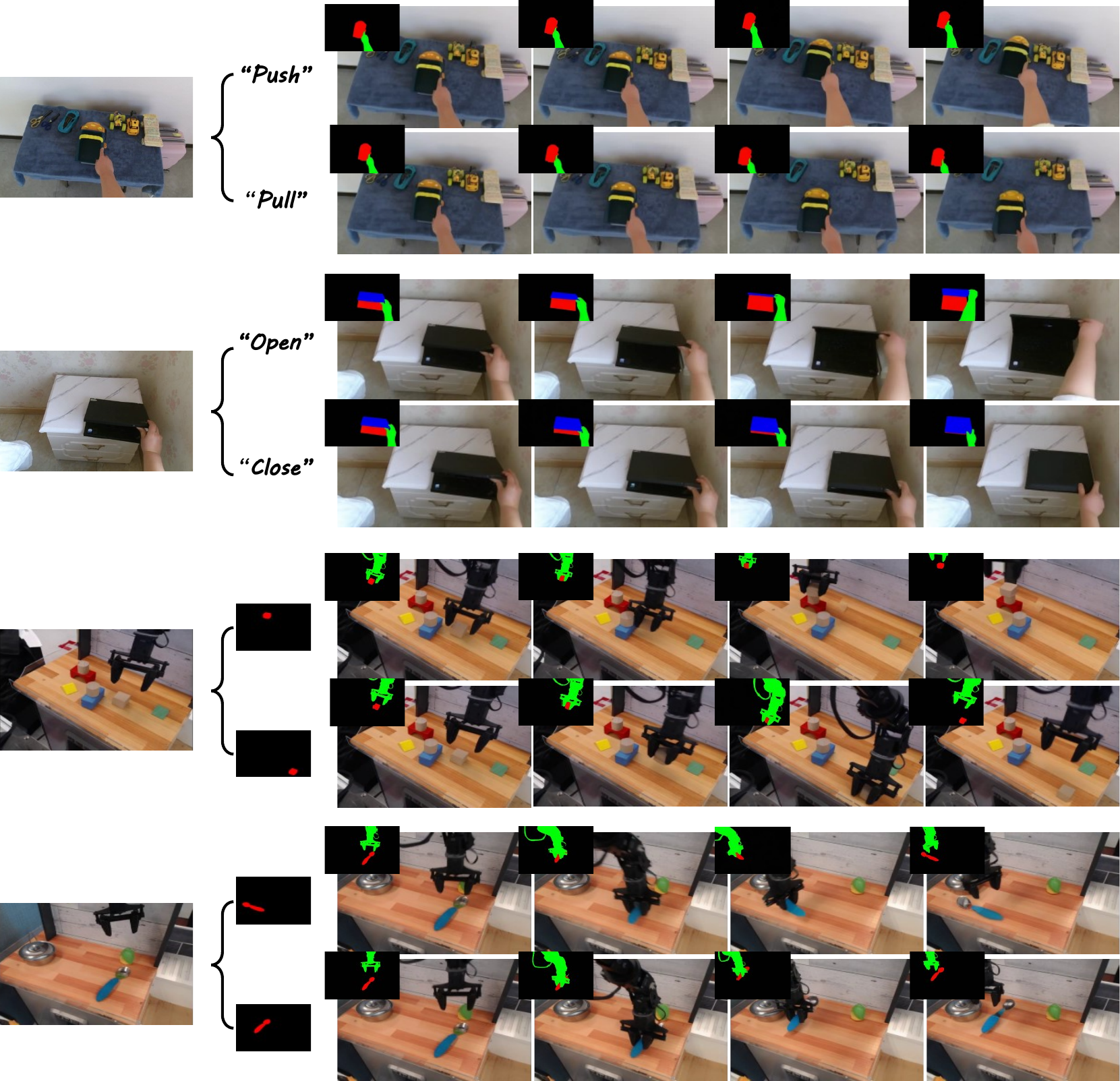}
  \caption{Qualitative analysis of different text prompts and position masks.}
  \label{fig:quali_analysis_text}
\end{figure*}

\begin{table}[t]
    \centering
    \begin{tabular}{lcccc}
      \toprule
       & \textbf{FVD$\downarrow$} & \textbf{LPIPS$\downarrow$} & \textbf{PSNR$\uparrow$} & \textbf{SSIM$\uparrow$}  \\
      \midrule
      ControlNet & 157.38 & 0.182 & 21.49 & 0.747 \\ 
      MaskLatent & 130.07 & 0.157 & 22.33 & 0.760 \\
      \midrule
      + object mask & 115.14 & 0.132 & 23.85 & 0.802 \\
      + random d/e & 108.80 & 0.124 & 24.16 & 0.802  \\
      + contact loss & 104.61 & 0.126 & 24.37 & 0.804 \\
      \bottomrule
    \end{tabular}
    \caption{Ablation results on the HOI4D dataset conducted with ground-truth mask trajectories. Random d/e denotes random dilation or erosion applied to the masks.}
    \label{tab:ablation}
\end{table}


\subsection{Quantitative Analysis}

Quantitative comparisons with baselines on HOI4D and BridgeData V2 are shown in~\cref{tab:comparison-hoi4d}.
We observe that the original pre-trained DynamiCrafter performs poorly on both datasets, with particularly degraded results on BridgeData V2. 
This highlights the limitations of current image-to-video generation models in capturing complex human-object or robot-object interactions.
After fine-tuning on each dataset, we see a notable improvement in performance, suggesting that domain adaptation helps the model better align with interaction-specific data distributions.
However, it lacks the ability to control the motion of actors and objects during the interaction process, as well as to specify the target object of interaction.
To ensure a fair comparison, we re-implement and adapt several control-based baselines under our experimental settings.
Across both datasets and multiple evaluation metrics, our method consistently outperforms all other baselines.
This performance gain demonstrates the effectiveness of our trajectory-conditioned design and joint modeling of actors and objects in capturing fine-grained, controllable interactions, while preserving video quality and temporal coherence.


\subsection{Qualitative Analysis}

In~\cref{fig:quali_analysi}, we present visual comparisons between our method and two representative baselines.
Our method consistently produces higher-quality interaction videos, characterized by coherent motion and clear contact regions.
In contrast, the baseline methods often suffer from visual artifacts, such as inconsistent color patterns (\textit{e.g.}, the white lid of a trashcan turning green) and incomplete object manipulation, where only parts of an object are affected (\textit{e.g.}, the handle and the pot appear detached).
Moreover, a notable advantage of our method lies in its flexible and versatile controllability. By simply modifying the input object mask, \m{} can synthesize diverse interaction sequences involving different objects.
This capability is illustrated in~\cref{fig:quali_analysis_obj}, where different target objects are grasped under the same input image, showcasing strong generalization to object variation.
We further analyze the controllability of our method under different conditioning prompts.
As shown in~\cref{fig:quali_analysis_text}, given the same input image, our framework allows users to adjust the text prompt or position mask to guide the generation process.
This enables fine-grained control over the type of interaction, such as pushing or pulling a toy car, or opening or closing a laptop.
It also supports spatial control over the object's placement, allowing the specification of both the location and orientation of the target object, which facilitates the generation of diverse and robust data for downstream robot learning tasks.


\subsection{Ablation Study}

An ablation study is performed in~\cref{tab:ablation} to evaluate the contribution of each component.
We first examine different control schemes and find that directly concatenating mask latent features with the noise input yields stronger performance than training an auxiliary ControlNet, leading to improved training stability and faster convergence in early stages.
After integrating object masks into the trajectory, we observe consistent improvements across all metrics.
These gains highlight the importance of modeling interactions through the motions of both actors and objects, rather than relying solely on actor representations.
The application of random mask perturbations further boosts performance by enhancing the model's robustness to variations in mask shape during training.
Lastly, the inclusion of the contact weighting loss leads to additional gains in video quality, particularly in accurately synthesizing interaction regions between the actor and the object.




\section{Conclusion}
In this paper, we introduce the task of interaction-centric video generation, focusing on synthesizing dynamic interactions between actors (humans or robots) and objects. We propose a novel two-stage pipeline that decouples interaction trajectories modeling from video synthesis, offering a more controllable and practical solution.
To support training and evaluation in this domain, we further introduce two dedicated benchmarks covering both human-object interaction and robotic manipulation scenarios.
Experimental results validate the effectiveness of our approach, highlighting its potential to advance controllable and realistic video generation in interactive settings.

\section{Acknowledgments}
This work is supported by a Start-up Grant from Nanyang Technological University and jointly funded by the Singapore Ministry of Education (MOE) under a Tier-1 research grant.

\bibliography{main}

\end{document}